\documentclass[sigconf]{acmart}

\usepackage{booktabs} 
\usepackage{hyperref}
\usepackage{url}
\usepackage{adjustbox}
\usepackage{makecell}

\setcopyright{rightsretained}

\acmDOI{10.475/123_4}

\acmISBN{123-4567-24-567/08/06}

\acmConference[KDD 2018]{ACM SIGKDD Conference}{Aug 2018}{London, UK}
\acmYear{2018}
\copyrightyear{2018}

\acmArticle{4}
\acmPrice{15.00}

\begin{document}
\title{SMILES2vec}
\subtitle{An Interpretable General-Purpose Deep Neural Network for Predicting Chemical Properties}

\author{Garrett B. Goh}
\affiliation{\institution{PNNL}}
\email{garrett.goh@pnnl.gov}

\author{Nathan Hodas}
\affiliation{\institution{PNNL}}
\email{nathan.hodas@pnnl.gov}

\author{Charles Siegel}
\affiliation{\institution{PNNL}}
\email{charles.siegel@pnnl.gov}

\author{Abhinav Vishnu}
\affiliation{\institution{PNNL}}
\email{abhinav.vishnu@pnnl.gov}

\renewcommand{\shortauthors}{Goh G.B. et al.}

\begin{abstract}
Chemical databases store information in text representations, and the SMILES format is a universal standard used in many cheminformatics software. Encoded in each SMILES string is structural information that can be used to predict complex chemical properties. In this work, we develop SMILES2vec, a deep RNN that automatically learns features from SMILES to predict chemical properties, without the need for additional explicit feature engineering. Using Bayesian optimization methods to tune the network architecture, we show that an optimized SMILES2vec model can serve as a general-purpose neural network for predicting distinct chemical properties including toxicity, activity, solubility and solvation energy, while also outperforming contemporary MLP neural networks that uses engineered features. Furthermore, we demonstrate proof-of-concept of interpretability by developing an explanation mask that localizes on the most important characters used in making a prediction. When tested on the solubility dataset, it identified specific parts of a chemical that is consistent with established first-principles knowledge with an accuracy of 88\%. Our work demonstrates that neural networks can learn technically accurate chemical concept and provide state-of-the-art accuracy, making interpretable deep neural networks a useful tool of relevance to the chemical industry.
\end{abstract}

%
%



\maketitle

\section{Introduction}

In the chemical industry, designing chemicals with desired characteristics is a bottleneck in the development of new products. Despite decades of research, much of modern day chemical design is still driven by serendipity and chemical intuition, although improvements to rational chemical design has been incrementally improving over time.~\cite{jorgensen2004} Currently, there exists two key contributing factors in the current state in physics-based or rule-based computational chemistry methods that accounts for the gap towards true rational chemical design. One factor is driven by technical limitations, such as that in compute capacity, and the other is incomplete or partial understanding of the underlying chemical concepts. The first factor is being addressed with development of GPU-accelerated code for molecular modeling~\cite{stone2010} and special-purpose chemistry supercomputers~\cite{shaw2014}, but the second factor requires solutions grounded in further fundamental research, which is often a slower process.

\subsection{Limitations of Feature Engineering and Black Box Models}
For the prediction of chemical properties that cannot be easily computed through physics-based or rule-based methods, modern \textit{in-silico} modeling in chemistry is therefore predicated on correlating engineered features with the activity or property of the chemical, which is formally known as the field of Quantitative Structure-Activity or Structure-Property Relationship (QSAR/QSPR) modeling~\cite{cherkasov2014}. Feature engineering in chemistry is a sophisticated science that stretches back to the late 1940s~\cite{platt1947}. Molecular descriptors, as they are termed by chemists, are basic computable properties or sophisticated descriptions of a chemical's structure, and these engineered features were developed based on first-principles knowledge. To date, over 5000 molecular descriptors have been developed~\cite{todeschini2008}. In addition, molecular fingerprints have also been designed, which instead of computing a basic property, provides a description of a specific part of the chemical's structure~\cite{rogers2010}. 

Since the 1980s, various machine learning (ML) algorithms have been used to predict the activity or property of chemicals~\cite{cherkasov2014} using molecular desciptors and/or fingerprints as input features. More recently, deep learning (DL) models have also been developed~\cite{dahl2014, mayr2016, ramsundar2015, hughes2016}. In general, these models either perform at parity or slightly outperform prior state-of-the-art models based on traditional ML algorithms for chemical applications~\cite{goh2017r}.

Compared to computer vision (CV) and natural language processing research (NLP), the use of DL models in chemistry relies heavily on these engineered features. This may be problematic as it limits the neural network's search space of potentially learnable representations. This is further exacerbated in situations in which engineered features are not appropriate or inadequate due to the lack of well-developed domain knowledge, which originates from the second factor that limits the impact of existing computational chemistry methods. In contrast, the dominant approach in CV/NLP research is to train DL models directly on typically unaltered raw data of large datasets with little or minimal feature engineering research. For example, unaltered images are used as the input in various CNN models~\cite{he2015} and similarly unaltered text is used in LSTM-based models~\cite{wu2016}. Therefore, developing DL models that leverage on representation learning is a logical advance for the field of chemistry as well.

Lastly, an additional challenge associated with current ML/DL models is the lack of interpretability. Typically operated as opaque black-box models, it is difficult to gain any scientific understanding as to \textit{why or how} an algorithm predicts a particular chemical property. For typical applications in CV/NLP research, this may not be an issue. However, for the chemical industry, particularly for regulated products, such as requesting FDA approval for new drugs, an explanation of how the chemical works (i.e. interacts with the body is a requirement). Therefore, models that have increased interpretability or explainability is not only of industrial relevance, but it will also enable chemists to formulate new hypothesis to improve on and possibly accelerate the pace fundamental research.

\subsection{Related Work}

In chemistry, raw unaltered data would typically refer to a representation that describes the structure and orientation of a chemical. In basic chemistry education, students are taught how to draw a 2D diagram of a chemical (i.e. an image), which also serves as the primary medium of communication amongst chemist. Alternatively, the same structural information can be encoded as graphs. Indeed, convolutional neural network (CNN) models that use chemical images~\cite{goh2017c1, goh2017c2, wallach2015} and other DL models that use molecular graphs~\cite{duvenaud2015, kearnes2016} have been recently developed. In addition, a chemical's structural information can also be encoded in text format, such as SMILES~\cite{weininger1988}, which is also the basis for interoperability between various cheminformatics software packages. In terms of text representations, we acknowledge there has been some prior work in this direction~\cite{jastrzkebski2016, bjerrum2017}. In terms of interpretable DL models, while we have seen advances in conventional CV/NLP applications~\cite{ribeiro2016}, at the time of writing, we are not aware of any interpretable DL models in chemistry that learns directly on raw data.

\subsection{Contributions}

Our work improves the existing state of learning directly from chemical text representations, and it is also the first interpretable neural network that works on chemical text. In the process, our work also addresses the following question: \textit{Is the SMILES representation sufficient to capture the first order distinction between different chemical properties?} Assuming that the above hypothesis is true, \textit{would it then be possible to validate what a neural network learns with established first principles knowledge on simple chemical properties, such as solubility?} Our contributions are as follows:
\begin{itemize}

	\item We perform extensive experiments to determine the optimal neural network architecture for interpreting the SMILES ``chemical language".
	\item We developed an explanation mask to explain \textit{why or how} the neural network makes a particular prediction.
	\item We show how an optimal SMILES2vec network architecture can be generalized to predict broad range of properties that are of relevance to multiple industries, including pharmaceuticals, biotechnology, materials and consumer goods.
	\item We demonstrate that SMILES2vec models, despite having no feature engineering, achieves better accuracy than contemporary multi-layer perceptron (MLP) models that uses engineered features.
\end{itemize}

The organization for the rest of the paper is as follows. In section 2, we examine the datasets, its broad applicability to chemical-affliated industries, as well as the Bayesian optimization process used for refining the network architecture, and the training protocols of the neural network. Then, in section 3, we document the experiments used to develop the final SMILES2vec network architecture and constructing of the interpretability masks. Lastly, in section 4, we perform additional experiments to quantify the accuracy of interpretability using the solubility dataset as an example, and evaluate SMILES2vec accuracy of chemical property predictions against contemporary DL models.

\section{Methods}

Here, we document the methods used in the development of SMILES2vec. First, we provide a brief introduction to SMILES. Then, we provide details on the datasets used, data splitting and preparation. Then, we examine the details of refining the neural network architecture using Bayesian methods, as well as the training protocol and evaluation metrics for the neural network.

\subsection{Introduction to SMILES}

SMILES is a ``chemical language"~\cite{weininger1988} that encodes structural information of a chemical into a compact text representation. There is a regular grammar to SMILES. For example, the alphabets denote atoms, and in some cases also what type of atoms. For example, c and C denote aromatic and aliphatic carbons respectively. Special characters like `=' denote the type of bonds (connections between atoms). Rings are denoted by encapsulating numbers, and side chains by round brackets. Thus, with sufficient training a chemist can read SMILES and infer the structure of the chemical. From this structural information, more complex properties can be predicted.

Inspired by language translation RNN work~\cite{wu2016}, we do not explicitly encode information about the grammar of SMILES. Instead, we expect that the RNN should learn these patterns and if necessary use them to develop intermediate features that would be relevant for predicting a variety of chemical properties.

\subsection{Datasets Used}

Our work creates a RNN model for general chemical property prediction, and ideally it should work effectively for different types of properties without significant network topology changes. To ensure that our results are comparable with contemporary DL models reported in the literature~\cite{wu2017} and earlier work on Chemception CNN models~\cite{goh2017c1, goh2017c2}, we used the Tox21, HIV, and FreeSolv dataset from the MoleculeNet benchmark~\cite{wu2017} for predicting toxicity, activity and solvation free energy respectively. These datasets (see Table~\ref{table:1}) represent a good mix of dataset sizes, type of chemical properties and regression vs classification tasks. In addition, we also used the ESOL solubility dataset to evaluate the interpretability of SMILES2vec.

\begin{table}[!t] 
		\begin{center}
		\begin{tabular}{|c|c|c|c|}
				\hline
				Dataset & Property & Task & Size \\
				\hline\hline
				Tox21 & \makecell{Non-Physical \\(Toxicity)} & \makecell{Multi-task \\ classification} & 8014 \\
				\hline
				HIV & \makecell{Non-Physical \\(Activity)} & \makecell{Single-task \\ classification} & 41,193 \\
				\hline
				FreeSolv & \makecell{Physical \\(Solvation)} & \makecell{Single-task \\ regression} & 643\\
				\hline
				ESOL & \makecell{Physical \\(Solubility)} & \makecell{Single-task \\ regression} & 1128 \\
				\hline
		\end{tabular} 
		\end{center}
\caption{Characteristics of the 4 datasets used to evaluate the performance of SMILES2vec.}
\label{table:1}
\end{table}

\subsection{Relevance to Chemical Industries}
In terms relevance to chemical-affliated industries, toxicity prediction has importance for chemicals that require FDA approval, which includes drugs and other therapeutics (pharmaceuticals) as well as cosmetics (consumer goods).~\cite{kruhlak2007} Activity is a measurement of how well a chemical binds to its intended target and is one of the factors that determine how well a chemical may perform as a drug.~\cite{buchwald2002} Therefore, accurate activity predictions are of relevance to both pharmaceuticals and biotechnology industries. Predicting solubility is an important consideration for developing formulations for products relevant to pharmaceuticals and consumer goods,~\cite{di2012} and it also affects the bioavailability of drugs. Lastly, free energy values itself are computable by physics-based simulations, and such methods are currently being employed by pharmaceuticals, consumer goods and materials industries.~\cite{chodera2011}

\subsection{Data Preparation}

The length of the SMILES string directly impacts the compute resources required to train RNN models. To maintain a balance between maximum amount of SMILES data, but also rapid training time, we surveyed the ChEMBL database, a collection of industrially-relevant chemicals that has over 1 million entries.~\cite{gaulton2011} Using this database as a proxy for relevant chemicals, we calculated that setting a maximum length of 250 characters would encompass 99.9\% of existing entries. Therefore, in the above-listed datasets, we excluded entries of more than 250 characters in the dataset.

Next, we created a dictionary that mapped the unique characters as one-hot encodings. Zero padding was also applied to ensure that shorter strings had a uniform size of 250 characters. In addition, extra padding of 10 zeroes were added both to the left and right of the string. Apart from the above-mentioned steps, no additional data augmentation steps were performed.

\subsection{Data Splitting}
We used a dataset splitting approach that is similar to that reported in previous work~\cite{goh2017c1}. A separate test set was first partitioned out to serve as a test for model generalizability. For the Tox21 and HIV dataset, 1/6th was partitioned out to form the test set, and for the FreeSolv and ESOL dataset, 1/10th was used to create the test set. The remaining 5/6th or 9/10th of the dataset was then used in the random 5-fold cross validation approach for training.

Model performance and early stopping criterion was determined by validation loss. Lastly, we oversampled the minority class for classification tasks (Tox21, HIV) to mitigate class imbalance. This was achieved by computing the ratio of both classes, and appending additional data from the smaller class by that ratio. The oversampling step was performed after stratification, to ensure that the same molecule is not repeated across training/validation/test sets.

\subsection{Bayesian Optimization of Neural Network Design}

We used a Bayesian optimizer, SigOpt~\cite{dewancker2016} to optimize the hyperparameters related to the neural network topology. Each different set of network hyperparameters are defined as a separate trial, and for each trial, we trained the model to completion using a standardized supervised training protocol. After each trial, the validation metric (AUC for classification tasks, RMSE for regression tasks) was used as input to the Bayesian optimizer for suggesting new network designs.

To prevent overfitting during this optimization process, the splitting of the dataset between training and validation sets was governed by a random seed. However, a fixed test set was maintained throughout, and this is also not used in the optimization process. By comparing the difference in validation and test set metrics, it would thus allow us to determine if the network design was being overfitted to the training/validation data. No hyperparameters optimization was performed for the learning protocol. Lastly, it should be noted that only a subset of the dataset was used in the Bayesian optimization. Specifically, we used only a single task (nr-ahr toxicity) from the Tox21 dataset and the Freesolv dataset.

\subsection{Training the Neural Network}

SMILES2vec was trained using a Tensorflow backend~\cite{abadi2016} with GPU acceleration using NVIDIA CuDL libraries.~\cite{chetlur2014} The network was created and executed using the Keras 2.0 functional API interface~\cite{chollet2015}. Thee RMSprop algorithm~\cite{hinton2012} was used to train for 250 epochs using the standard settings recommended (learning rate = $10^{-3}$, $\rho = 0.9$, $\epsilon = 10^{-8}$). The batch size was 32, and we also included early stopping to reduce overfitting. This was done by monitoring the loss of the validation set, and if there was no improvement in the validation loss after 25 epochs, the last best model was saved as the final model.

For classification tasks, we used the binary crossentropy loss function for training. The performance metric reported in our work is area under the ROC curve (AUC). For regression tasks, we used the mean average error as the loss function for training. The performance metric reported is RMSE. Unless specified otherwise, the reported results in our work denote the mean value of the performance metric, obtained from the 5 runs in the 5-fold cross validation. 

\section{Experiments}

In this section, we first conduct several Bayesian optimization experiments to optimize SMILES2vec's architecture and hyperparameters. Then, we conduct further experiments to develop an explanation mask for improving interpretability of the model. 

\subsection{SMILES2vec Neural Network Design}

RNNs, particularly those based on LSTMs~\cite{hochreiter1997} or GRUs~\cite{cho2014} are effective neural network designs for learning from text data. Its effectiveness has been demonstrated in examples like the Google Neural Translation Machine that uses an architecture of 8+8 layers of residual LSTM units~\cite{wu2016}. Most of the other reported application of RNNs in NLP research are similarly used to model sequence-to-sequence predictions, and often fewer (i.e. 2 to 4) layers have been found to be sufficiently accurate for their tasks.

Our work differs from conventional NLP research as we are modeling sequence-to-vector predictions, where the sequence is a SMILES string, and the vector is a measured chemical property. Because of this, and also because SMILES is a fundamentally different language, commonly-used techniques in NLP research, such as embeddings like Word2vec ~\cite{mikolov2013} cannot be easily adapted for use in our work. Therefore, a substantial component of our work is in the design of the RNN architecture specific to SMILES.

\subsection{Architectural Class Exploration}

\begin{figure}[!htbp]
\centering
\includegraphics[scale=0.35]{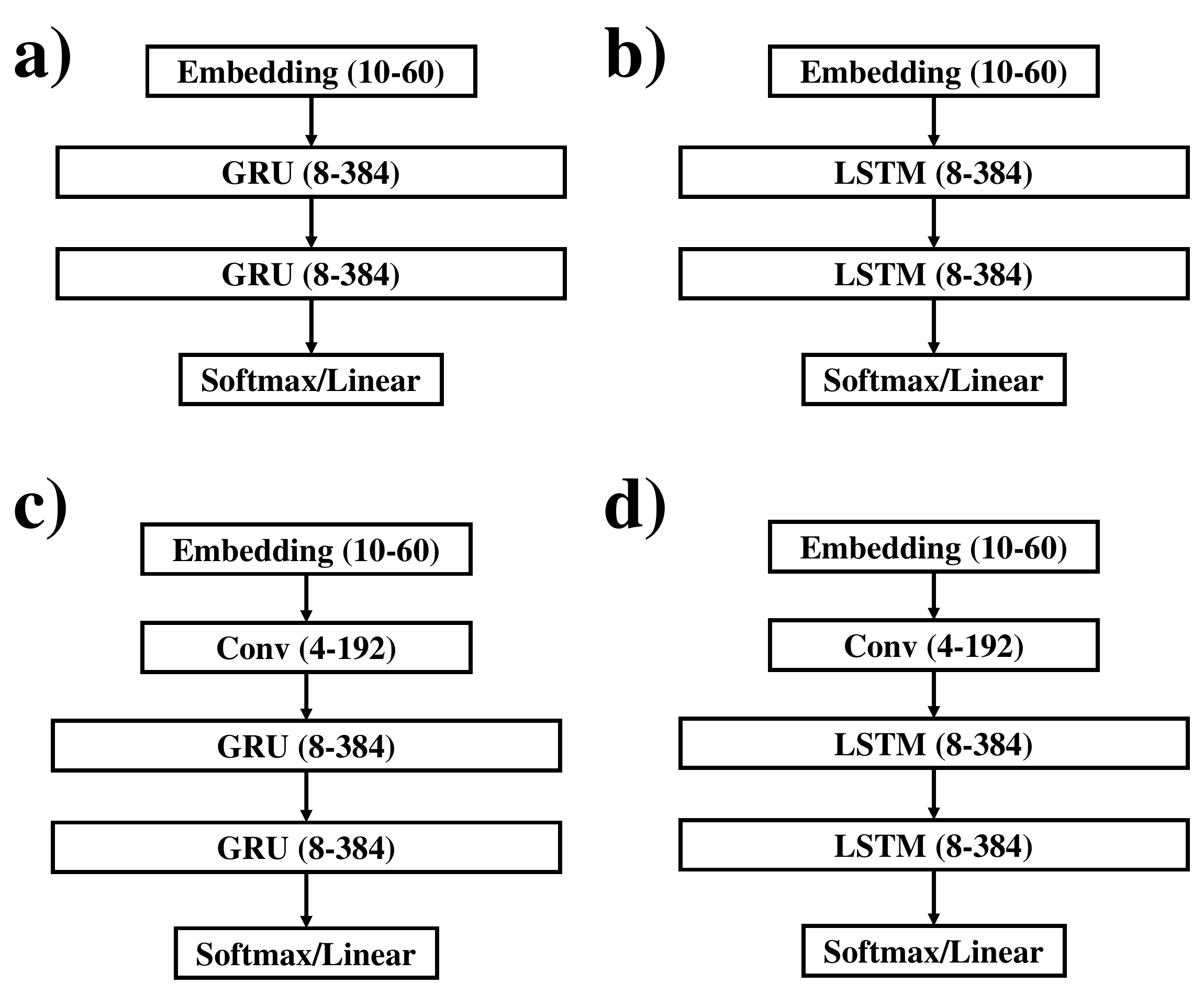}
\caption{\small Illustration of the 4 architectural classes investigated, (a) GRU, (b) LSTM, (c) CNN-GRU and (d) CNN-LSTM. Number of units explored is indicated in parenthesis.}
\label{fig:1}
\end{figure}

We first explore the RNN model's architecture class, which primarily includes high-level design choices, such as the type of units used, type of layers, arrangement of layers, etc. LSTMs and GRUs are the two major RNN units used in the literature, and form the basis of two architectural classes. The template design for each  class starts with an embedding layer that feeds into a 2-layer bidirectional GRU or 2-layer bidirectional LSTM as illustrated in Figure~\ref{fig:1}. In addition, we explored the utility of adding a 1D convolutional layer between the embedding and GRU/LSTM layers. This design forms the template of the other two architectural classes explored.

A separate Bayesian optimization was used to optimize the hyperparameters of each architectural class. Specifically, we varied the size of the embedding from 10 to 60 in intervals of 10. The number of units in the GRU/LSTM layers ranged from 8 to 384 in intervals of 8, and the number of units in the convolutional layer ranged from 4 to 192 in intervals of 4. For the convolutional layer, a size of 3 and a stride of 1 was used, which is based on the design principles from modern convolutional neural network~\cite{szegedy2015}. No additional optimization was performed on the size or stride of the convolutional layers. In addition, no specific shape of the network topology was enforced.

\subsection{Bayesian Optimization of Hyperparameters}

In order for Bayesian methods to be effective, a sufficient number of trials for different neural network design has to be performed. In practice, it has been recommended that a minimum of 10N trials be performed, where N is the number of tunable hyperparameters. In our work, we performed 60 trials for each of the 4 architectural class. In addition, we manually seeded 6 initial designs for each class. Specifically, we used initial designs that had an embedding size of 40, a convolution layer with 16 filters, and both LSTM/RNN layer with [8, 16, 32, 64, 128, 256] units.

In addition, because we are developing a general-purpose neural network design that can be re-used for a broad range of property prediction, it would not be feasible to include all conceivable training data to optimize the network design within the limits of available computing resources. Therefore, a subset of the datasets were used in the Bayesian optimization (see section 2.5 for details), and separate optimizations were performed for the Tox21 classification and the FreeSolv regression tasks.

\begin{figure}[!htbp]
\centering
\includegraphics[scale=0.5]{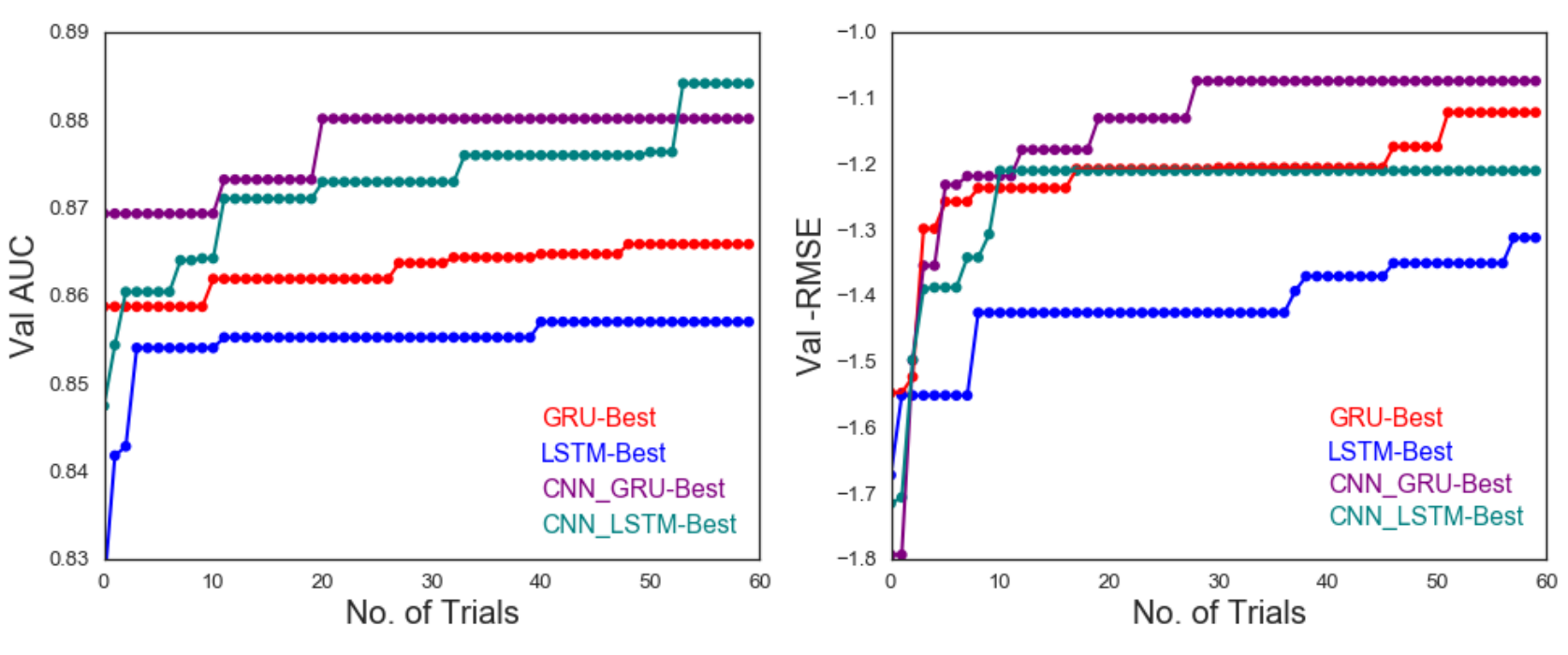}
\caption{\small Results of Bayesian optimization of the hyperparameters of the 4 architectural classes for the (a) Tox21 classification and (b) Freesolv regression tasks.}
\label{fig:2}
\end{figure}

The results of the Bayesian optimization across all 4 classes and 2 tasks are as indicated in Figure~\ref{fig:2}. For Tox21 classification, we observed that an additional convolutional layer between the embedding and RNN/LSTM layers improved model performance relative to their counterparts, and the best performing model was the CNN-LSTM class, with CNN-GRU trailing slightly behind. For FreeSolv regression, we observed that GRU-based networks outperform LSTM-based networks. Taking into considerations for generalization to other type of chemical properties, we selected the CNN-GRU architectural class for the remainder of this work. Then, we selected the best network design of this class, which is summarized in Table~\ref{table:2}.

\begin{table}[!t] 
		\centering
		\begin{adjustbox}{max width=\columnwidth}
		\begin{tabular}{|c|c|c|c|}
				\hline
				em\_size & \#conv & \#rnn1 & \#rnn2 \\
				\hline
				50 & 192 & 224 & 384 \\
				\hline
		\end{tabular} 
		\end{adjustbox}
\caption{Best CNN-GRU network design for the final SMILES2vec model.}
\label{table:2}
\end{table}

\begin{figure}[!htbp]
\centering
\includegraphics[scale=0.5]{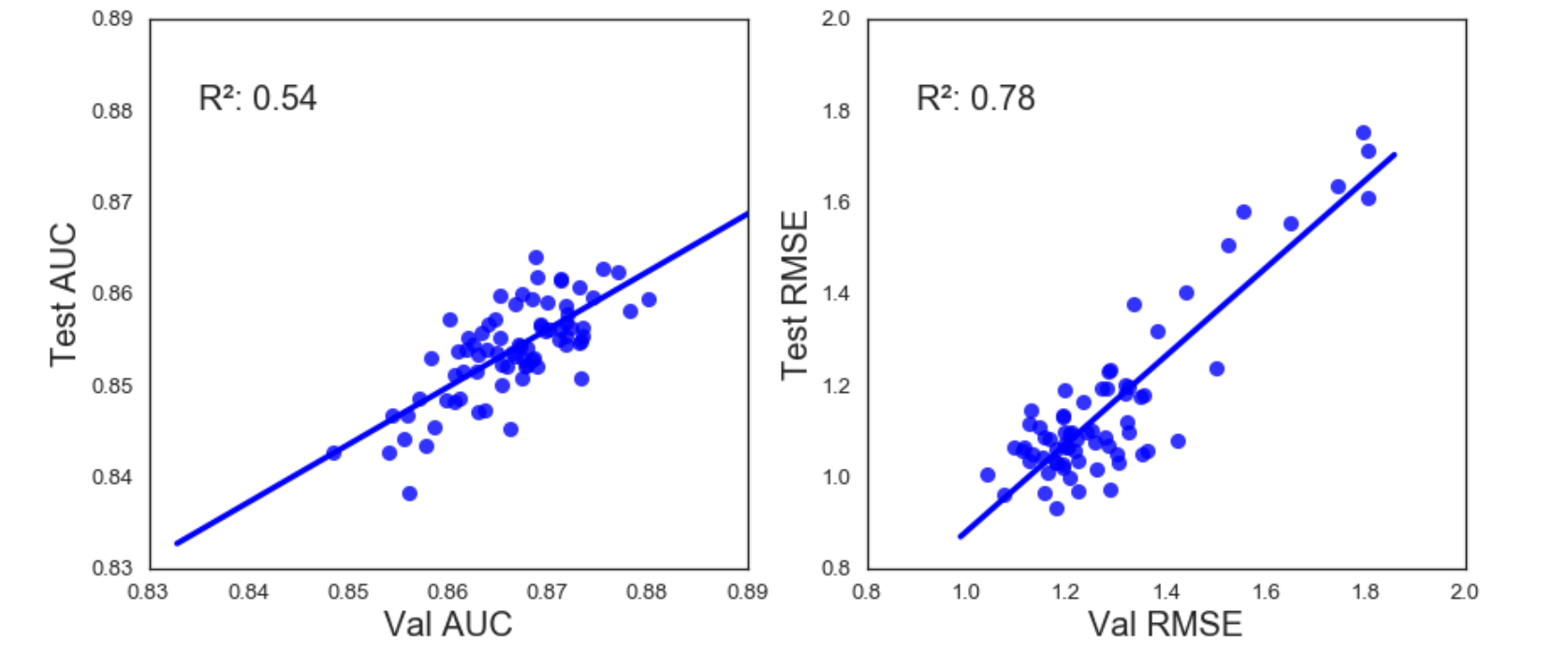}
\caption{\small Correlation plot between validation and test metrics for all trials enumerated for the (a) Tox21 and (b) Freesolv datasets.}
\label{fig:3}
\end{figure}

Lastly, because the Bayesian algorithm uses the validation metric as a means to optimize the network's hyperparameters, there is a possibility that as one progresses, there may be overfitting towards the validation set. To determine the extent of overfitting, we examined the correlation between the validation metrics (whose validation set data would be changing during the Bayesian optimization) and the test metrics (whose test set data is fixed, and was never used in the Bayesian optimization). As illustrated in Figure~\ref{fig:3}, the correlation between validation and test metrics is 0.54 for the Tox21 dataset and 0.78 for the FreeSolv dataset. The lower correlation of the Tox21 dataset relative to the FreeSolv dataset may be explained by noting the AUC performance metric on which the optimization was performed, is not the same as the crossentropy loss function used for training the network.

\subsection{SMILES2vec Interpretation}
To gain a better insight into the SMILES2vec model, we developed a method to gain some level of interpretability. Here, our objective is to identify the part(s) of the SMILES string that is responsible for the neural network's decision. 

Methods for  explaining ``black box" models exist~\cite{ribeiro2016}, but most of these methods tend to require explicit combinatorial analysis.  The approach we provide here  provides insight into how the neural network analyzes the data, without combinatorially probing the input. This is achieved by training an explanation mask, whereby a separate explanation network learns to mask input data to produce near identical output as would be obtained from the original data.

\subsection{Training the Explanation Mask}

\begin{figure}[!htbp]
\centering
\includegraphics[scale=0.5]{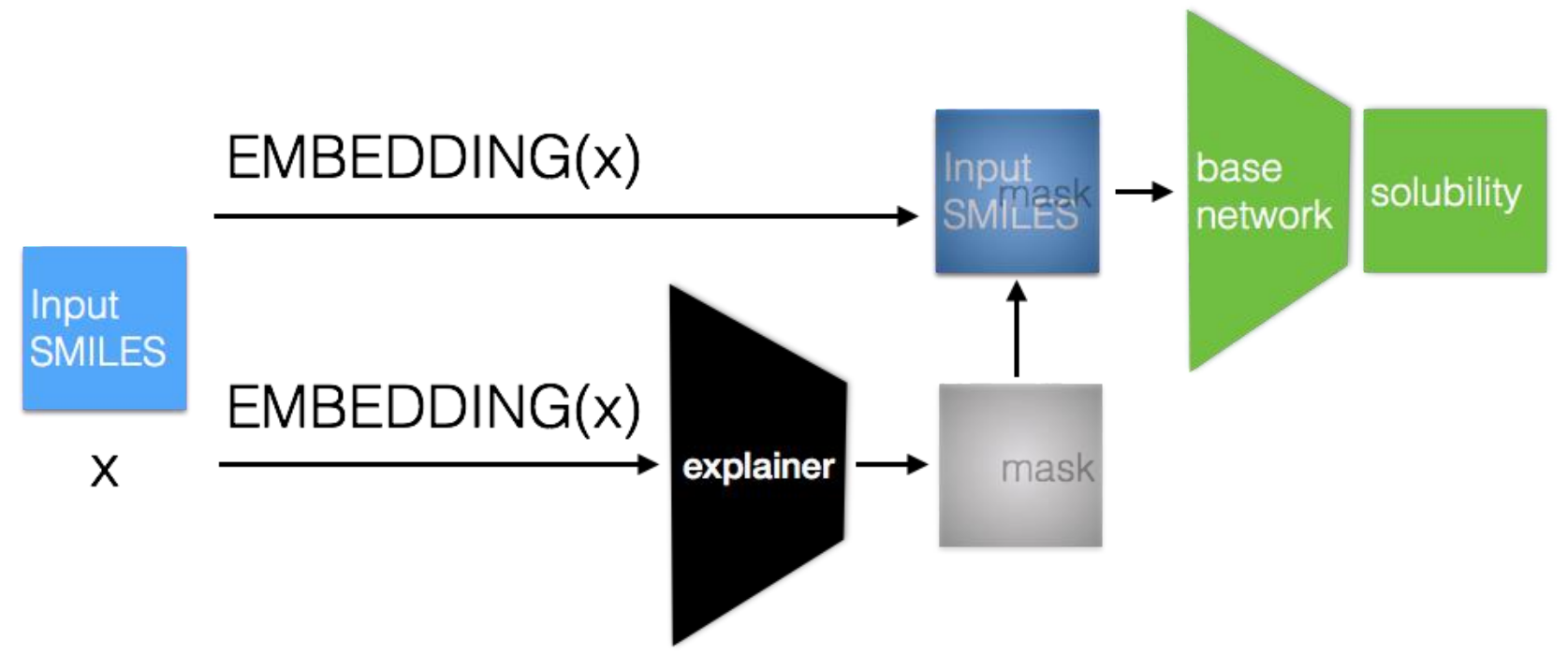}
\caption{\small Structure for training the explanation network. The SMILES input is passed through the embedding layer, then into the explainer. This produces a mask that is placed over the original embedding and sent through the pre-trained base model.}
\label{fig:4}
\end{figure}

We train a neural network generated mask to identify the important characters of the input. The procedure is as follows: First, we use the final SMILES2vec model (Table~\ref{table:2}) as the base network. Next, we construct another neural network to produce a mask over the input data, with the objective to train the mask such that the output of the base neural network remains the same but it masks as much data as possible. We freeze the base neural network, and we train the explanation mask end-to-end, as shown in Figure~\ref{fig:4}.

With the weights of the base network frozen, the mask being learned will be specific to the SMILES2vec model.  Each input will produce a different mask. To avoid the mask being trivial (completely uniform), we added two forms of regularization, a small L2 regularization, and we also penalized the mask for having high entropy (putting equal weight on all inputs). The overall loss function for a single element of each mini-batch is as follows:
\begin{align*}
Loss_i = &||f(SMILES_i,\theta) - Sol(SMILES_i)||_2  \\
		&+ 1e^{-6}||MASK_i||_2 + 0.05 H(MASK_i),
\end{align*}
where $f(SMILES_i,\theta)$ is the base neural network applied to the ith SMILES, $Sol(SMILES_i)$ is the solubility, $H$ is the entropy over the mask (normalized to sum to 1), and $MASK_i$ is the vector of the calculated explanation mask at each entry in the input SMILES string.
	
The explanation network used to create the mask was a 20 layer residual network with SELU~\cite{klambauer2017} activations. The padding was fixed such that the length of the input remained the same at each layer. The input to the network is the embedding of the SMILES string. The last layer is a 1D convolution of length 1, followed by a batch normalization, then by a softplus activation. We observed that the final batch normalization layer to be very important for trainability. This provides an output between 0 and infinity at each SMILES position.   A mask output of 0 would prevent  the base SMILES2vec from receiving that input character.  A mask of $\infty$ would cause the SMILES2vec to put more attention on that input character. We trained with Adam~\cite{kingma2014} until convergence. We started the learning rate at $10^{-2}$ and divided by 10 as the training error plateaued, ultimately training down to $10^{-6}$.

\section{Performance}

In this section, we quantity the accuracy of interpretability on the solubility dataset, Then, we demonstrate generalizability of the SMILES2vec model by evaluating its performance on other datasets.

\subsection{Interpreting Chemical Solubility}

We demonstrate proof-of-concept for an interpretable SMILES2vec network using the ESOL solubility dataset.~\cite{wu2017} Chemical solubility is a well-understood and simple chemical property where there is established first-principles knowledge. Briefly, parts of a chemical (i.e. functional groups) can typically be classified as either hydrophilic or hydrophobic. Hydrophilic ``water-loving" groups, like alcohols, amines and carboxyl form strong interactions with water and increase the overall solubility of a compound, and they typically contain non-carbon atoms like nitrogen and oxygen. The reverse is true for hydrophobic ``water-hating" groups, which tend to make chemicals more insoluble, and they typically are carbon-based chains/rings and halogens (chlorine, bromine, iodine).

We  used the pre-trained the SMILES2vec base model, which attained a validation RMSE of 0.63. Solubility values are reported on the log10 scale, with less soluble compounds having more negative number. The mask outputs a normalized attention value that denotes the importance of a particular character in the network's decision. For each SMILES string, we identified the top-3 characters (see Table~\ref{table:3} for examples). Then, we separated the dataset into soluble (> -1.0) and insoluble (< -5.0) compounds. Using established knowledge of chemical solubility to establish the ground truth, we expect that soluble compounds should have higher attention on the atoms O, N, and insoluble compounds to have higher attention on atoms C, F, Cl, Br, I. With this ground truth labeling in expected atoms, we computed the top-3 accuracy of SMILES2vec interpretability, which is 88\%.
	
\begin{figure}[!htbp]
\centering
\includegraphics[scale=0.5]{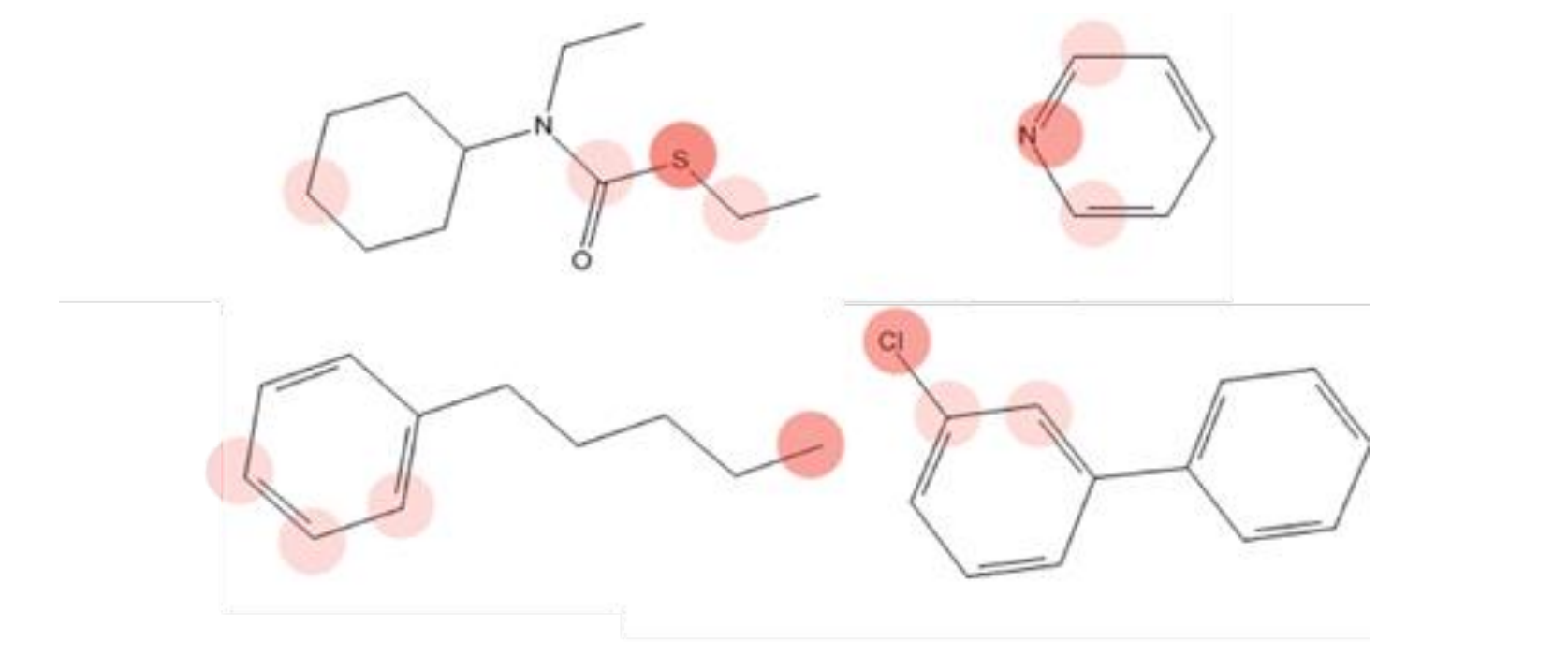}
\caption{\small Colored circles of increasing darkness indicate the locations of increasing attention on the molecule. The explanation mask validates established knowledge by focusing on atoms of known hydrophobic and hydrophilic functional groups.}
\label{fig:5}
\end{figure}

In addition, we also qualitatively examined the outputs of the masks by mapping the SMILES character to the corresponding atom(s) in the molecular structure, and examples are shown in Figure~\ref{fig:5}. For molecules with low solubility, the characters c, C, and Cl tend to receive more attention than others, which correspond to hydrophobic groups. In contrast, molecules with high solubility have attention focused on the characters O and N, which correspond to hydrophilic groups.

\begin{table}[!t] 
		\centering
		\begin{adjustbox}{max width=\columnwidth}
		\begin{tabular}{|c|c|c|}
				\hline
				SMILES & Solubility & Top-3 Chars \\
				\hline
				c1\textbf{ccn}cc1 & 1.18 & c,c,n \\
				O=\textbf{Cc}1ccc\textbf{o}1& -0.87 & C,c,o \\
				Clc1c\textbf{c}c(cc1Cl)c2cc\textbf{cc}c2 & -5.93 & c,c,c \\
				C\textbf{c}1\textbf{c}2\textbf{c}cccc2c(C)c3ccccc13 & -6.91 & c,c,c\\
				\hline
		\end{tabular} 
		\end{adjustbox}
\caption{Sample SMILES entries, their predicted solubility value, and the top-3 most important characters are bolded.}
\label{table:3}
\end{table}

The localization of appropriate atoms on each functional group type depending on the chemical's predicted solubility value indicates that SMILES2vec has learned representations that correspond to known chemistry concepts. Lastly, we emphasize that SMILES2vec has developed these representations without being provided any explicit chemical information. While chemical information is implicitly encoded in the SMILES string, no ``decoding solution" was provided to the network, neither was further feature engineering required. Our work therefore demonstrates the effectiveness of representation learning from raw data in the chemical sciences. 

\subsection{Generalization of SMILES2vec Models}

Thus far, we have only evaluated the performance of SMILES2vec on the ESOL solubility dataset. Unlike solubility, the other 3 datasets (toxicity, activity, solvation energy) are more complex properties, for which no simple rule-based methods exist in the chemistry literature. Without the ability to generate ground-truth labels, quantifying the accuracy of the SMILES2vec interpretation is non-trivial, and is beyond the scope of this work. Nevertheless, the accuracy of the model's predictions can still be evaluated.

We also note that the architectural optimization of SMILES2vec only included a small fraction of the 4 datasets identified for this work; the HIV dataset was not included, and 11 out of 12 toxicity tasks were not included. Hence, this section also determines how generalizable the Bayesian optimized network design will be to other chemical tasks.

First, we determined the effectiveness of generalizing SMILES2vec to the 3 remaining datasets. The following validation performance metrics were obtained: AUC of 0.80 for the full Tox21 dataset, AUC of 0.78 for the HIV dataset, and RMSE 1.4 kcal/mol for the FreeSolv dataset. Furthermore, we note that in all models, the difference between the validation and test metrics is small, further confirming the generalization of the model to compounds it has not seen either during the model training, or in the Bayesian hyperparameter optimization. Using a recently developed pre-training approach,~\cite{goh2017c3} we were also able to improve the performance of SMILES2vec slightly, attaining AUC of 0.81 for the full Tox21 dataset, AUC of 0.80 for the HIV dataset, and RMSE 1.2 kcal/mol for the FreeSolv dataset. 

Based on these results, we conclude that the Bayesian optimization of the network architectural design was effective in developing a general-purpose SMILES2vec network design for other chemical properties. We also note in recent literature there has been a trend towards using other "black box" approaches as a solution for network architecture design, for example using RNNs and reinforcement learning to optimize the design of a target neural network.~\cite{zoph2016} However, such methods typically require on the order of \textasciitilde10,000 trials, which is much more than the \textasciitilde500 trials used in our work. In addition, given that the template of each architectural class was fixed, adaptive methods that automtically grow or shrink the neural network are also viable alternatives to network design.~\cite{siegel2016}

Next, we compare the performance of the best SMILES2vec model against contemporary deep neural networks that have reported results on the same datasets (Tox21, HIV, FreeSolv) that we have evaluated our model on. We compare against a typical MLP network that uses engineered features~\cite{wu2017}, a chemistry-specific molecular graph convolutional neural network~\cite{wu2017}, and Chemception, a deep CNN that uses images~\cite{goh2017c2}.

\begin{figure}[!htbp]
\centering
\includegraphics[scale=0.38]{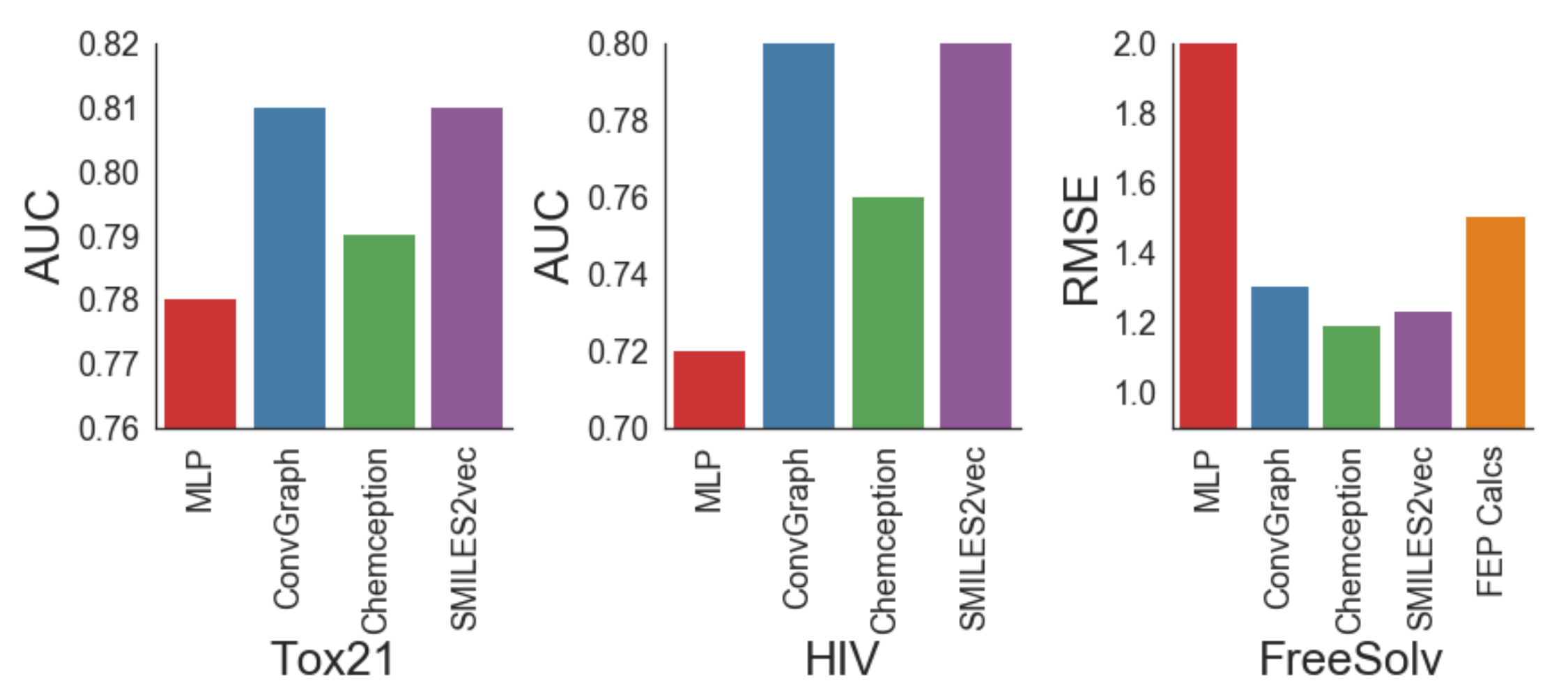}
\caption{\small Performance of SMILES2vec against contemporary deep neural networks trained on engineered features, image and graph data. For Tox21 and HIV, higher AUC is better. For FreeSolv, lower RMSE is better.}
\label{fig:6}
\end{figure}

The results are presented in Figure~\ref{fig:6}, and we use validation metrics to evaluate the quality of the model. In comparing the 4 methods, we observed that the standard MLP DL models that uses engineered features (fingerprints) performed the worst. SMILES2vec outperformed CNN models such as Chemception in classification tasks, but slightly underperformed in regression tasks. As indicated in our previous work, it is likely that the lack of atomic number information, which is not embedded in the SMILES format is responsible for its lower performance in predicting calculable physical properties.~\cite{goh2017c2} In addition, SMILES2vec also outperformed first-principles models for computing solvation free energy (note: there are no first-principles models for computing toxicity/activity), which is especially noteworthy since SMILES2vec (and neural networks in general) can predict values much faster than traditional computational chemistry simulations, which typically require minutes to hours for each calculation. Against convolutional graphs, which is the current state-of-the-art for many chemical tasks, SMILES2vec either matches for classification tasks (Tox21: 0.81 vs 0.81, HIV: 0.80 vs 0.80) or outperforms for regression task (FreeSolv: 1.2 vs 1.3). Therefore, SMILES2vec is not only as accurate as the current state-of-the-art in chemistry DL models, but more importantly it is also an interpretable model.

\section{Conclusion}

In this paper, we develop SMILES2vec, the first general-purpose deep neural network that uses chemical text data (SMILES) for predicting chemical property, with an explanation mask that improves interpretability. By performing extensive Bayesian optimization experiments, we identified a specific CNN-GRU neural network architecture is effective in predicting a wide range of properties. SMILES2vec achieved a validation AUC of 0.81 and 0.80 for Tox21 toxicity and HIV activity prediction respectively, and a validation RMSE of 1.2 kcal/mol and 0.63 for solvation energy and solubility. Using the solubility dataset as an illustration of SMILES2vec interpretability, we construct explanation masks that indicate SMILES2vec localizes on specific characters in hydrophilic or hydrophobic groups, with a top-3 accuracy of 88\%. Identification of such functional groups and their relationship to chemical solubility is a key first-principles concept in chemistry, which SMILES2vec was able to discover on its own. Compared to other DL models, SMILES2vec's accuracy outperforms the typical MLP DL models that uses engineered features as input. Against the current state-of-the-art (convolutional graph networks), SMILES2vec outperforms on regression tasks and matches on classification tasks. These results indicate that SMILES2vec can accurately predict a broad range of properties and learn technically accurate chemical concepts, which suggest that it can be used as an interpretable tool for the future of deep learning driven chemical design.

\begin{acks}
The authors would like to thank Dr. Nathan Baker for helpful discussions. This work is supported by the following PNNL LDRD programs: Pauling Postdoctoral Fellowship and Deep Learning for Scientific Discovery Agile Investment.
\end{acks}

\bibliographystyle{ACM-Reference-Format}
\bibliography{sample-bibliography}

\end{document}